\documentclass{article}

 \usepackage[main, final]{neurips_2026}

\usepackage[utf8]{inputenc} 
\usepackage[T1]{fontenc}    
\usepackage{hyperref}       
\usepackage{url}            
\usepackage{booktabs}       
\usepackage{amsfonts}       
\usepackage{nicefrac}       
\usepackage{microtype}      
\usepackage{xcolor}         
\usepackage{xspace}         

\usepackage{graphicx}
\usepackage{multirow, makecell}
\usepackage{bbding}
\usepackage{pifont}
\usepackage{caption}
\usepackage{subcaption}
\usepackage{wrapfig}
\usepackage{enumitem}
\usepackage{amsmath}
\usepackage{amssymb}
\usepackage[misc]{ifsym}

\newcommand{\vs}{\textit{vs}.}

\newcommand{\method}{Next Forcing\xspace}

\title{Next Forcing: World Action Modeling with Multi-Chunk Prediction}

\author{
    Gangwei Xu$^{1,2}$ \quad
    Qihang Zhang$^{1,\dagger}$ \quad 
    Jiaming Zhou$^{1,4}$ \quad
    Xing Zhu$^1$ \quad 
    \\ [4pt]
    \textbf{Yujun Shen$^1$ \quad
    Xin Yang$^{2,5,\ddagger}$ \quad
    Yinghao Xu$^{3,1,\ddagger}$}
    \\ [6pt]
    {\normalfont $^1$Robbyant \quad $^2$HUST \quad $^3$HKUST \quad $^4$HKUST (GZ) \quad $^5$Optics Valley Laboratory}
    \\ [6pt]
    {\normalfont $^{\dagger}$Project Lead \quad $^{\ddagger}$Corresponding Author}
}

\begin{document}

\maketitle

\begin{abstract}
  Autoregressive video generation has emerged as a powerful paradigm for World Action Models (WAMs). However, existing approaches suffer from slow training convergence and limited converged accuracy, particularly at high frame rates, as the training supervision is confined to the current chunk without explicit signals about future dynamics; they also suffer from slow inference due to iterative video denoising. In this paper, we present \method, a multi-chunk prediction (MCP) framework for world action modeling that enables faster training, higher accuracy, and accelerated inference. Inspired by multi-token prediction in large language models, \method introduces an MCP training objective that augments the main model with lightweight auxiliary MCP modules to simultaneously denoise video chunks at multiple future temporal horizons (next$^1$, next$^2$, next$^3$ chunks). These MCP modules form a causal chain across prediction depths, where intermediate features fused from multiple layers of the main model are leveraged to predict future dynamics, allowing near-future predictions to inform farther-future ones and providing dense multi-scale temporal supervision back to the main model. During training, the MCP modules significantly accelerate convergence and improve converged accuracy, especially at high frame rates: at 50 fps, \method achieves a 93.1\% relative improvement over LingBot-VA at 5k training steps and 2.3$\times$ faster convergence, and establishes new state-of-the-art results on the RoboTwin benchmark (94.1/93.5\% on Clean/Random). At inference, the MCP modules can be retained to predict the next video chunk in parallel with the current one, achieving 2$\times$ inference acceleration. Beyond World Action Models, \method also improves physical-law adherence on PhyWorld and reduces FVD by over 50\% in action-free general video pretraining. Project website: \url{https://gangweix.github.io/next-forcing/}.

\end{abstract}

\section{Introduction}
\label{introduction}

\begin{figure}[!t]
    \centering
    \includegraphics[width=\linewidth]{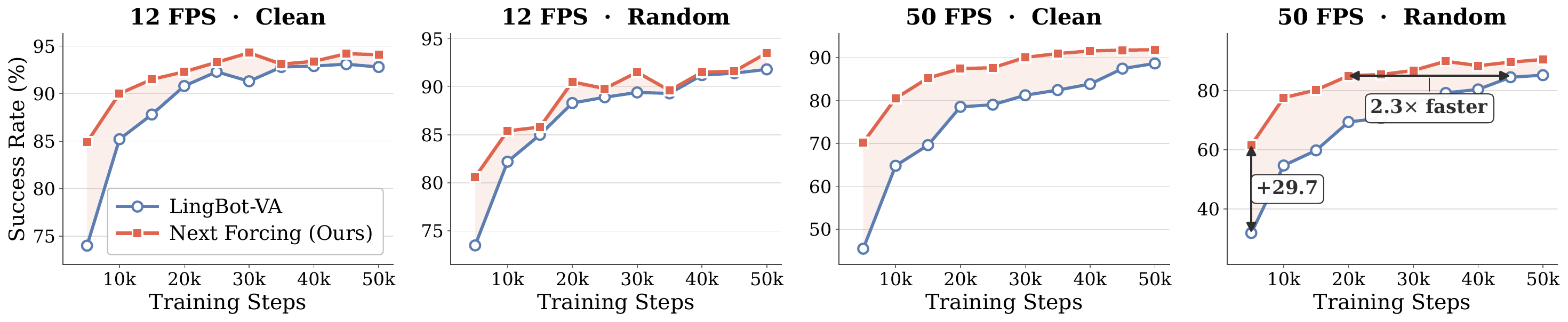}
    \caption{\textbf{Task success rate (\%) on RoboTwin across training steps.} \method converges faster and reaches higher final accuracy than LingBot-VA at both 12 and 50\,fps. The advantage is most pronounced at 50\,fps: at 5k steps \method already outperforms LingBot-VA by 29.7 points on Random, and matches its 45k-step accuracy at only 20k steps, a 2.3$\times$ training speedup.}
    \label{fig:teaser}
\end{figure}

Videos capture how the physical world evolves and how agents act within it, recording the dynamics of physical interactions at scale~\cite{ha2018world, dreamerv3, gaia1, gaia2, genie, diamond, vpt}.
Building on this, World Action Models (WAMs)~\cite{lingbot, dreamzero} have recently emerged as a new paradigm for embodied AI, learning manipulation policies by jointly modeling future video and actions.
The dominant training objective for WAMs is \emph{teacher-forced next-chunk denoising}, where the model denoises the noisy current chunk conditioned on clean past chunks from ground-truth data.
Despite the recent achievements of this paradigm \cite{lingbot, dreamzero}, teacher-forced next-chunk denoising remains an inefficient and shortcut-prone training signal for video world models.
More precisely, predicting the next chunk is a fundamentally \emph{local} task, which admits an appearance shortcut~\cite{shortcut}: since adjacent chunks are visually highly similar, much of the denoising loss can be driven down by learning a near-identity map from the clean past chunk to the current one, with only small residual corrections. Such a shortcut is much easier to fit than the underlying dynamics and can absorb a large share of the gradient signal, leaving weak pressure on the model to capture the long-range temporal evolution that governs how the scene actually changes.
We refer to this as \emph{myopic supervision}, and its cost becomes particularly acute at high frame rates: at 50\,fps, the appearance gap between adjacent chunks narrows to the point where the shortcut is nearly lossless, and standard teacher forcing converges significantly slower with lower final accuracy (Figure~\ref{fig:teaser}).

Our central insight is that turning the local single-chunk objective into a \emph{long-range multi-chunk} objective forces the model to learn the latent dynamics governing temporal evolution rather than relying on appearance shortcuts.
This idea has been validated in language modeling, where multi-token prediction (MTP) \cite{deepseekv3, gloeckle2024better} trains auxiliary modules to predict multiple future tokens, improving sample efficiency and enabling inference acceleration.
Adapting it to video world models, however, is non-trivial: prediction targets are \emph{continuous} video latents rather than discrete language tokens, generation proceeds through \emph{iterative denoising} rather than single-step sampling, and temporal dependencies span \emph{multiple horizons} of varying scale.

We present \textbf{\method}, a multi-chunk prediction (MCP) framework that addresses myopic supervision for better and faster world action modeling.
We use \emph{world action modeling} to refer to jointly modeling future visual dynamics and robot actions within a unified generative architecture.
\method augments the main model with a small set of auxiliary MCP modules that simultaneously predict video chunks at multiple future horizons (next$^1$, next$^2$, next$^3$) and form a causal chain across prediction depths.
To let the temporal supervision propagate deep into the main model's representations, the MCP modules fuse intermediate features from \emph{multiple layers} of the main model rather than the final layer alone.
The MCP modules are trained with a \emph{higher timestep shift} than the main model, forcing them to rely more heavily on the main model's representations and tightening the coupling between the two. 
Although the MCP objective directly supervises the video stream, the improved future representations propagate to the action stream through cross-modal attention.

Our work joins a growing family of ``forcing'' methods for autoregressive video generation.
Existing methods vary either \emph{what context the model sees} (as in teacher forcing \cite{lingbot, dreamzero} and self forcing \cite{selfforcing}) or \emph{how noise is scheduled} (as in diffusion forcing \cite{diffusionforcing}).
\method is orthogonal to both: it varies \emph{what the model is asked to predict}, and is therefore composable with all of the above.

We primarily evaluate \method as a World Action Model on RoboTwin~\cite{robotwin}. We further evaluate its action-free form on PhyWorld~\cite{phyworld} and general video pretraining to verify that its temporal supervision benefits extend beyond robot action learning. \method sets new state-of-the-art on RoboTwin (94.1/93.5\% on Clean/Random), achieves 2.3$\times$ faster convergence at 50 fps (Figure~\ref{fig:teaser}), 2$\times$ inference acceleration, substantial gains on PhyWorld, and over 50\% FVD reduction on general video pretraining with 3.5M in-house video clips, confirming its effectiveness beyond robot-specific data. Our main contributions are as follows:


\begin{itemize}
\item We propose \textbf{\method}, a multi-chunk prediction framework for World Action Models that addresses the myopic supervision of autoregressive video prediction.
\item \method establishes new state-of-the-art results on the RoboTwin benchmark with significantly faster training convergence. At 50 fps, \method achieves 2.3$\times$ faster convergence and higher converged accuracy.
\item Beyond improving training, the MCP modules can be retained at inference to predict the next video chunk in parallel with the current one, accelerating generation.
\item We provide systematic ablations on the design choices that enable multi-chunk prediction to transfer from discrete tokens to continuous video latents (timestep shift, multi-layer feature fusion, and causal MCP chain), offering practical guidance for future work (Section~\ref{experiments}).
\end{itemize}

We hope \method motivates further investigation of training \emph{objectives} (beyond context construction and noise scheduling) in autoregressive video generation.

\begin{figure}[t]
    \centering
    \includegraphics[width=1\linewidth]{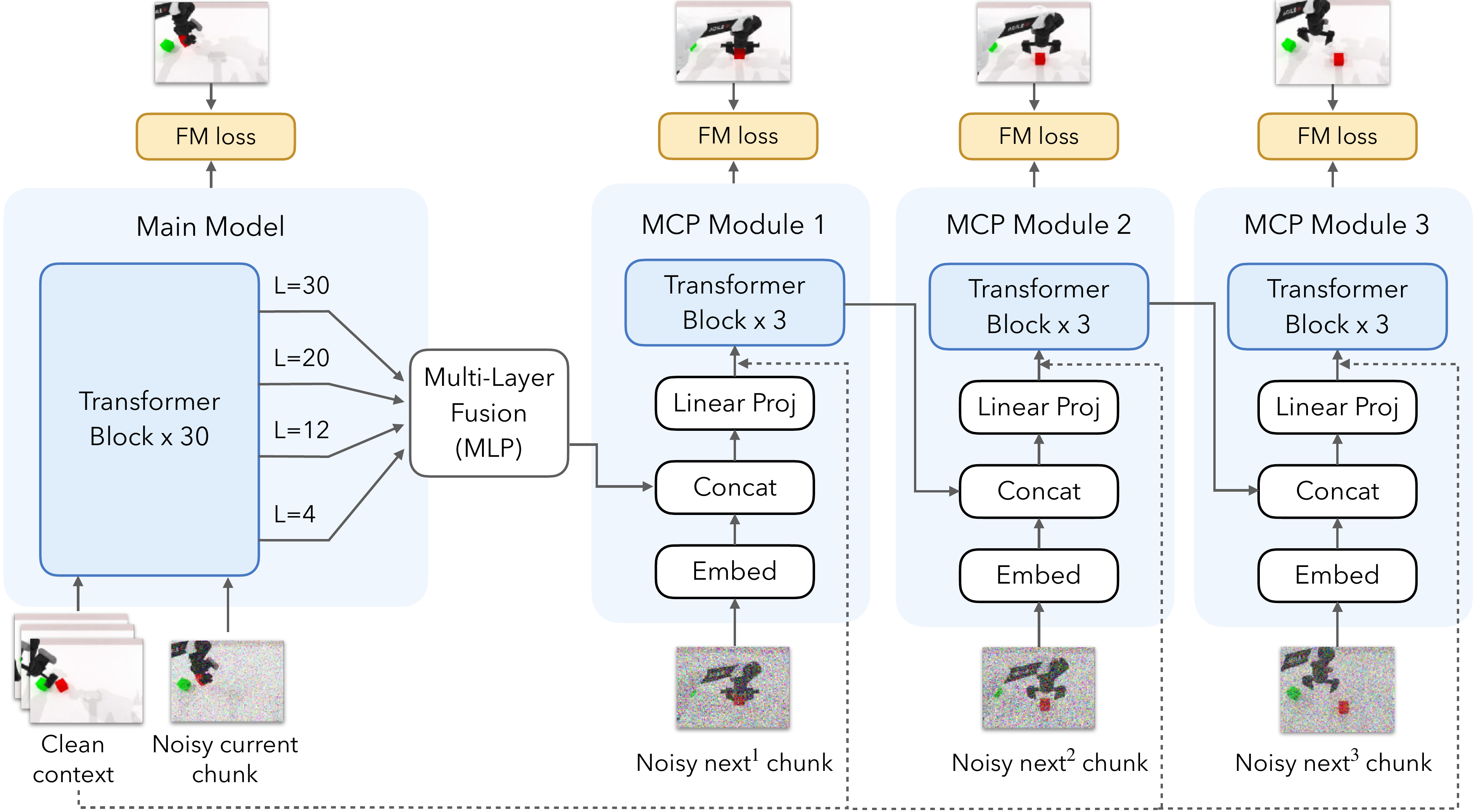}
    \caption{\textbf{Overview of Next Forcing.} The main model denoises the current chunk, while chained MCP modules predict future chunks (next$^1$, next$^2$, $\ldots$) using features from the
   main model, providing dense temporal supervision during training and enabling parallel chunk prediction at inference.}
    \label{fig:architecture}
\end{figure}

\section{Related Work}
\label{related_work}

\subsection{World Action Models}
Unlike Vision-Language-Action (VLA) models \cite{rt2, openvla, xvla, pi0, pi05, pi07, gemini, rdt, being, diffusionpolicy, act, liu2026stamo, tracevla, cotvla, spatialvla, tinyvla, groot, magma, hpt, zhou2026exploring} that directly map visual observations and language instructions to actions, World Action Models (WAMs) \cite{dreamzero, lingbot, dreamgen, shen2025videovla, cen2025worldvla, gr1, gr2, vla3d,beingh07} incorporate video prediction into robot policy learning, first predicting future visual dynamics and then decoding robot actions from the predicted frames. Several recent works have advanced the WAM paradigm along different axes. DreamZero \cite{dreamzero} demonstrates that a 14B-parameter autoregressive video diffusion model, trained on diverse non-repetitive robot data, achieves zero-shot generalization to novel tasks and cross-embodiment transfer. LingBot-VA \cite{lingbot} proposes a unified autoregressive framework with teacher forcing that jointly learns video prediction and action execution, achieving state-of-the-art bimanual manipulation through closed-loop control with persistent KV cache memory. Concurrently, Fast-WAM \cite{fastwam} finds that the primary benefit of video modeling lies in improving world representations during training. Other works explore joint video-action generation from complementary perspectives, including latent action representations\cite{motus, chen2025moto, lapa, lapo}, joint video-action diffusion \cite{uwm,won2025dual}, video-conditioned policy learning \cite{unipi, vpp, gen2act, liang2024dreamitate, tian2024predictive, avdc}, and video generation as policy \cite{cosmospolicy, mimicvideo}. Our work is complementary to these efforts: rather than proposing a new WAM architecture or exploring data scaling, we focus on improving the \emph{training paradigm} of autoregressive WAMs through multi-chunk prediction, which is applicable to existing WAM frameworks and also enables inference acceleration.

\subsection{Autoregressive Video Generation}
Autoregressive video generation \cite{jin2024pyramidal,yin2025slow,liu2025rolling,chen2026out,ren2025next} has become the core generation paradigm for action-conditioned world models and world action models \cite{lingbot, dreamzero, cosmospolicy,zhang2026sparse}. The standard training approach is teacher forcing \cite{lingbot, dreamzero}, where the model learns to denoise the current chunk conditioned on clean ground-truth context. While stable, it creates a distribution gap between training (clean context) and inference (self-generated context), known as exposure bias~\cite{ranzato2016sequence, scheduledsampling, selfforcing}. Diffusion Forcing \cite{diffusionforcing} mitigates this by training with independently sampled noise levels per frame, so that the model encounters noisy context during training. Self Forcing \cite{selfforcing} takes a more direct approach by conditioning on self-generated histories during training, explicitly bridging the train-test distribution gap through a distribution-matching loss. These methods primarily address \emph{how context is constructed} during training or inference. In contrast, our \method addresses a different and complementary aspect: \emph{what the model is trained to predict}. By extending the prediction target from the current chunk to multiple next chunks, \method provides dense multi-scale temporal supervision that encourages trajectory-level temporal reasoning and avoids myopic supervision.

\section{Preliminaries}
\label{sec:prelim}

\subsection{Flow Matching}

Flow matching~\cite{lipman2023flow} is a generative modeling framework that learns a velocity field to transport samples from a noise distribution to the data distribution. Given a clean sample $\mathbf{x}_0$ and Gaussian noise $\boldsymbol{\epsilon} \sim \mathcal{N}(0, \mathbf{I})$, the noisy sample at time $t \in [0, 1]$ is constructed via linear interpolation:
\begin{equation}
    \mathbf{x}_t = (1 - t) \, \mathbf{x}_0 + t \, \boldsymbol{\epsilon}.
\end{equation}
A neural network $v_\theta(\mathbf{x}_t, t, \mathbf{c})$ is trained to predict the velocity $\mathbf{v}^* = \boldsymbol{\epsilon} - \mathbf{x}_0$ conditioned on context $\mathbf{c}$, with the training objective:
\begin{equation}
    \mathcal{L}_\text{FM} = \mathbb{E}_{t, \mathbf{x}_0, \boldsymbol{\epsilon}} \left[ \left\| v_\theta(\mathbf{x}_t, t, \mathbf{c}) - (\boldsymbol{\epsilon} - \mathbf{x}_0) \right\|^2 \right].
    \label{eq:fm_loss}
\end{equation}
The timestep $t$ is sampled with a timestep shift parameter $s$ (see Appendix~\ref{app:timestep_shift} for the detailed formulation). At inference, clean samples are generated by integrating the velocity field from $t=1$ (pure noise) to $t=0$ (clean data) using an ODE solver.

\subsection{Autoregressive Video Generation with Teacher Forcing}

We build upon the autoregressive video-action framework of LingBot-VA~\cite{lingbot}. The model operates on video latents encoded by a pre-trained VAE and generates video frames in chunks of $M$ frames. At each autoregressive step $i$, the model denoises the noisy current chunk $\mathbf{x}_t^{(i)}$ conditioned on the clean previous chunks $\mathbf{x}_0^{(1:i-1)}$ from ground-truth data and a language instruction $\ell$:
\begin{equation}
    v_\theta\!\left(\mathbf{x}_t^{(i)}, t, \left[\mathbf{x}_0^{(1:i-1)}, \ell\right]\right) \approx \boldsymbol{\epsilon}^{(i)} - \mathbf{x}_0^{(i)}.
    \label{eq:teacher_forcing}
\end{equation}
This teacher forcing formulation naturally aligns with closed-loop deployment, where ground-truth observations replace generated frames after each action execution. However, as discussed in Section~\ref{introduction}, the model's supervision is confined to the current chunk, leading to the myopic supervision problem.

\section{\method}
\label{method}

\subsection{Overview}

\method augments the standard teacher-forcing training objective with multi-chunk prediction (MCP), where auxiliary modules predict video chunks at multiple next temporal horizons in addition to the current chunk. Figure~\ref{fig:architecture} illustrates the overall framework. During training, the main model denoises the current chunk as in standard teacher forcing, while three auxiliary MCP modules denoise the next$^1$, next$^2$, and next$^3$ chunks through a causal chain, where each depth builds on the output of the previous one. At inference time, the MCP modules can either be discarded for zero-overhead deployment or retained for parallel chunk generation.
Our primary instantiation is a World Action Model that jointly predicts future visual dynamics and robot actions.

\subsection{Multi-Chunk Prediction Objective}
\label{sec:mcp_objective}

The key idea of \method is to extend the prediction target from the current chunk to multiple next chunks, providing explicit supervision about future dynamics. This encourages trajectory-level temporal reasoning during training and directly addresses the myopic supervision problem.

\paragraph{Temporal Chunk Shifting.}
Given a training video latent $\mathbf{x}_0 \in \mathbb{R}^{C \times F \times H \times W}$, where $C$, $F$, $H$, $W$ denote the channel, number of chunks, height, and width dimensions respectively. Each chunk contains $M$ frames, where $M$ is randomly sampled from $\{1, \ldots, M_\text{max}\}$ at each training step for robustness across temporal scales. Let $i$ denote the current chunk index. For each MCP depth $k \in \{1, 2, 3\}$, we construct a temporally shifted target $\mathbf{x}_0^{[k]}$ by advancing the video latent by $k$ chunks:
\begin{equation}
    \mathbf{x}_0^{[k]}[i] = \mathbf{x}_0\!\left[\min(i + k, \, F)\right],
    \label{eq:shift}
\end{equation}
where chunks beyond the sequence boundary are padded by replicating the last chunk. Each $\mathbf{x}_0^{[k]}$ is the video shifted $k$ chunks into the future.  

\paragraph{Independent Noise Injection.}
Each shifted target is independently noised using the flow matching formulation with its own timestep and noise sample:
\begin{equation}
    \mathbf{x}_{t_k}^{[k]} = (1 - t_k) \, \mathbf{x}_0^{[k]} + t_k \, \boldsymbol{\epsilon}_k, \quad \boldsymbol{\epsilon}_k \sim \mathcal{N}(0, \mathbf{I}),
\end{equation}
where $t_k$ is sampled with a dedicated timestep shift parameter $s_\text{mcp}$. We set $s_\text{mcp} > s_\text{main}$, which biases the MCP modules toward higher noise levels. The motivation is to strengthen the coupling between the MCP modules and the main model: at higher noise levels the MCP input carries less information about its own target, so the modules are forced to rely more heavily on the main model's representations to denoise. This pushes the MCP loss gradients into the main model rather than letting the lightweight MCP modules absorb the supervision themselves.

\paragraph{MCP Position Encoding.}
To inform the MCP modules of their temporal offset, the chunk shift is incorporated into the rotary position embeddings (RoPE)~\cite{roformer}:
\begin{equation}
    \text{RoPE}(\mathbf{x}_0^{[k]}[i]) = \text{RoPE}(i + k),
\end{equation}
so that each MCP module receives positional information for the future chunk it predicts.

\subsection{Chained MCP Modules}
\label{sec:architecture}

The MCP modules are designed to encourage the main model to develop trajectory-level temporal representations. To achieve this, we fuse intermediate features from multiple layers of the main model rather than only the final output, since early layers tend to capture coarse structural patterns while later layers refine fine-grained details. This multi-layer fusion allows the MCP supervision signal to influence the main model's representations at various depths through gradient backpropagation.

\paragraph{Multi-Layer Feature Fusion.}
During the main model's forward pass through its 30 transformer layers, we collect hidden states at 4 intermediate layers $\{4, 12, 20, 30\}$, capturing representations at various depths of the backbone. Notably, the collected hidden states include both the noisy current latent and the clean history latent, allowing the fusion to capture both the denoising state and the ground-truth context. These multi-scale features are concatenated along the feature dimension and compressed through a two-layer MLP:
\begin{equation}
    \mathbf{h}_\text{fuse} = \text{MLP}\!\left(\left[\mathbf{h}_{4}; \mathbf{h}_{12}; \mathbf{h}_{20}; \mathbf{h}_{30}\right]\right) \in \mathbb{R}^{B \times N \times d},
    \label{eq:fuse}
\end{equation}
where $N$ is the number of latent tokens and $d$ is the hidden dimension. During backpropagation, the MCP loss gradients flow through $\mathbf{h}_\text{fuse}$ back into these intermediate layers, providing temporally-aware supervision to both the early and late stages of the main model.

\paragraph{Causal Chain Across Depths.}
The three MCP modules form a causal chain. For each depth $k$, the noisy shifted target $\mathbf{x}_{t_k}^{[k]}$ is embedded through the shared patch embedding layer and fused with the output from the previous depth:
\begin{equation}
    \mathbf{z}^{[k]} = W_k \left[\mathbf{h}_\text{prev}^{[k-1]}; \, \text{Embed}(\mathbf{x}_{t_k}^{[k]})\right], \quad W_k \in \mathbb{R}^{d \times 2d},
    \label{eq:fuse2}
\end{equation}
where $\mathbf{h}_\text{prev}^{[0]} = \mathbf{h}_\text{fuse}$. The fused representation is processed through 3 lightweight transformer blocks to predict the flow matching velocity $\hat{\mathbf{v}}^{[k]}$. The output simultaneously serves as $\mathbf{h}_\text{prev}^{[k]}$ for the next depth, allowing depth-2 predictions to build on depth-1 features, and depth-3 on depth-2. The MCP modules share the same attention mask as the main model, requiring only a single mask construction per training step for efficient training (see Appendix~\ref{app:attn_mask} for details).

\subsection{Joint Video-Action Architecture}
\label{sec:video_action}

Following LingBot-VA~\cite{lingbot}, \method jointly models video prediction and action decoding. The joint prediction is decomposed into two stages: (1) predicting future visual dynamics, and (2) decoding actions via inverse dynamics:
\begin{equation}
    \mathbf{x}_{i+1} \sim p_\theta(\cdot \mid \mathbf{x}_{\leq i}, \mathbf{a}_{<i}, \ell), \quad
    \mathbf{a}_{i} \sim g_\psi(\cdot \mid \mathbf{x}_{\leq i+1}, \mathbf{a}_{<i}, \ell),
    \label{eq:wam}
\end{equation}
where $\mathbf{x}_{\leq i}$ denotes the observation history up to chunk $i$, $\mathbf{a}_{<i}$ denotes the action history, and $\ell$ is the language instruction. The video stream first predicts the next visual state $\mathbf{x}_{i+1}$, then the action stream decodes $\mathbf{a}_i$ by conditioning on the observations including the predicted future chunk $\mathbf{x}_{i+1}$.

Both streams are implemented within a unified Mixture-of-Transformers (MoT)~\cite{mot} architecture, where a video stream and an action stream fuse through cross-modal attention at each transformer layer. The MCP modules introduced in Section~\ref{sec:architecture} operate on the video stream. The improved video representations propagate to the action stream through the shared cross-modal attention, benefiting action decoding indirectly.
Therefore, although the MCP losses are directly applied to the video stream, their temporally enriched representations improve joint world-action learning and action decoding through cross-modal interaction.

\subsection{Training Objective}
\label{sec:training}

\paragraph{Main Loss.}
The main model is trained with two flow matching losses for the video and action streams. The video dynamics loss supervises the video stream to predict future visual states:
\begin{equation}
    \mathcal{L}_\text{video} = \mathbb{E}_{t, \mathbf{x}_0, \boldsymbol{\epsilon}} \left[ \left\| v_\theta(\mathbf{x}_t, t, \mathbf{c}) - (\boldsymbol{\epsilon} - \mathbf{x}_0) \right\|^2 \right],
\end{equation}
and the action loss supervises the action stream via inverse dynamics:
\begin{equation}
    \mathcal{L}_\text{action} = \mathbb{E}_{t, \mathbf{a}_0, \boldsymbol{\epsilon}} \left[ \left\| v_\psi(\mathbf{a}_t, t, \mathbf{c}_a) - (\boldsymbol{\epsilon} - \mathbf{a}_0) \right\|^2 \right],
\end{equation}
where $\mathbf{a}_0$ denotes the ground-truth action sequence and $\mathbf{c}_a$ includes the visual context from both current and future observations.

\paragraph{MCP Loss.}
Each MCP module is supervised with a flow matching loss on its temporally shifted target $\mathbf{x}_0^{[k]}$ (Eq.~\ref{eq:shift}). For depth $k$:
\begin{equation}
    \mathcal{L}_k^\text{MCP} = \mathbb{E}_{t_k, \mathbf{x}_0^{[k]}, \boldsymbol{\epsilon}_k} \left[ \left\| v_\theta^{[k]}(\mathbf{x}_{t_k}^{[k]}, t_k, \mathbf{c}) - (\boldsymbol{\epsilon}_k - \mathbf{x}_0^{[k]}) \right\|^2 \right],
    \label{eq:mcp_loss}
\end{equation}
where the last $k$ padded chunks are excluded from the loss computation.

\paragraph{Total Loss.}
The complete training objective combines the main losses and the MCP losses:
\begin{equation}
    \mathcal{L} = \mathcal{L}_\text{video} + \mathcal{L}_\text{action} + \sum_{k=1}^{3} w_k \cdot \mathcal{L}_k^\text{MCP}.
    \label{eq:total_loss}
\end{equation}

\subsection{Inference}
\label{sec:inference}

The MCP modules introduced in Section~\ref{sec:architecture} serve a dual role: they are trained as a supervision mechanism but can also be repurposed at inference. We support two inference modes that share the same trained checkpoint and trade off cost against throughput.

\paragraph{Zero-Overhead Mode.}
All MCP modules (the fusion MLP, projection layers, and lightweight transformer blocks) are discarded. The main model operates exactly as in the standard autoregressive pipeline: at each step, it denoises the current chunk, with the same architecture, latency, and memory footprint as the baseline. All quality gains in this mode come from the enriched training signal that the MCP objective propagates back into the main model during training, with no test-time cost.

\paragraph{Parallel Chunk Generation Mode.}

The MCP modules can also be retained at inference to amortize the cost of one main-model forward pass over multiple chunks, in spirit similar to speculative and parallel decoding for LLMs~\cite{speculative, medusa}. Concretely, in a single denoising trajectory the main model produces the current chunk while the depth-$1$ MCP module simultaneously produces the next chunk, and the MCP transformer blocks are an order of magnitude lighter than the main model, so adding them to the forward pass is nearly free. Each autoregressive step therefore advances the video by two chunks instead of one, yielding $2\times$ inference acceleration. The depth-2 and depth-3 MCP modules are not used in this mode, as their predictions are superseded by the main model in the next step, but the same mechanism could be extended for higher speedups at the cost of accumulated drift; we leave this exploration to future work.

The two modes use the \emph{same} trained model, allowing deployments to choose freely between strict baseline parity and $2\times$ throughput without retraining.

\section{Experiments}
\label{experiments}

\begin{table}[t]
  \caption{Evaluation on the RoboTwin benchmark (average success rate \% over 50 tasks). Clean: fixed initial configurations. Random: randomized object poses and scene layouts. Best results are \textbf{bolded}.}
  \label{tab:benchmark}
  \centering
  \small
  \begin{tabular}{@{}lcccccccc@{}}
    \toprule
    & X-VLA  & $\pi_0$ & $\pi_{0.5}$  & Motus & Being-H0.7  & Fast-WAM & LingBot-VA  & \textbf{\method} \\
    \midrule
    Clean  & 72.9 & 65.9 & 82.7 & 88.7 & 90.2 & 91.9 & 92.9 & \textbf{94.1} \\
    Random & 72.8 & 58.4 & 76.8 & 87.0 & 89.6 & 91.8 & 91.5 & \textbf{93.5} \\
    \bottomrule
  \end{tabular}
\end{table}

\begin{table}[t]
  \caption{Ablation studies on RoboTwin Clean subset, trained for 20k steps on 16 GPUs. Default baseline: $s_\text{main}$=5 with noisy history augmentation. Default MCP: $s_\text{mcp}$=10, multi-layer fusion, weight init from main model, 3 transformer blocks per depth.}
  \label{tab:ablation}
  \centering
  \small
  \begin{tabular}{@{}lc c lc@{}}
    \toprule
    \multicolumn{2}{c}{Baseline (LingBot-VA~\cite{lingbot}) Ablation} & & \multicolumn{2}{c}{MCP Module Ablation} \\
    \cmidrule(lr){1-2} \cmidrule(lr){4-5}
    Configuration & SR (\%) & & Configuration & SR (\%) \\
    \midrule
    Baseline (default) & 75.6 & & Baseline + MCP (default) & 85.8 \\
    \midrule
    $s_\text{main}$=1 & 65.3 & & $s_\text{mcp}$=5 & 83.2 \\
    $s_\text{main}$=10 & 78.4 & & w/o multi-layer fusion & 83.6 \\
    $s_\text{main}$=20 & 77.6 & & w/o weight init & 83.8 \\
    $s_\text{main}$=25 & 77.2 & & transformer blocks=1 & 86.5 \\
    w/o noisy history aug. & 69.8 & & transformer blocks=5 & 85.0 \\
    \bottomrule
  \end{tabular}
\end{table}

\subsection{Experimental Setup}

\paragraph{Benchmarks.}
We evaluate \method on two benchmarks. \textbf{RoboTwin}~\cite{robotwin} is a simulation benchmark containing 50 bimanual manipulation tasks that require coordinated dual-arm control. Each task is evaluated under two settings: \emph{Clean} (fixed initial configurations) and \emph{Random} (randomized object poses and scene layouts). We report the average task success rate over all 50 tasks. RoboTwin serves as our primary evaluation of world action modeling, where we measure closed-loop manipulation performance using task success rate. \textbf{PhyWorld}~\cite{phyworld} evaluates whether video generation models can discover and adhere to physical laws (e.g., uniform linear motion, elastic collision, parabolic motion) from visual data alone.

\paragraph{Implementation Details.}                                                  \method is built on the LingBot-VA framework with a Wan2.2~\cite{wan2025} Transformer backbone (30 layers). We keep all baseline settings identical to LingBot-VA. The main model uses a timestep shift of $s_\text{main} = 5$ with noisy history augmentation applied at probability 0.5. For the MCP modules, we use 3 prediction depths (next$^1$, next$^2$, next$^3$), each with 3 transformer blocks. The MCP timestep shift is set to $s_\text{mcp} = 10$. MCP loss weights are $w_1 = 0.5$, $w_2 = 0.2$, $w_3 = 0.1$. The chunk size $M$ is randomly sampled from $\{1, \ldots, M_\text{max}\}$ with $M_\text{max} = 4$. MCP module weights are initialized from the last few layers of the main model. Following LingBot-VA, we first pre-train on a large-scale multi-embodiment dataset and then post-train on RoboTwin. We adopt the same training setup as LingBot-VA~\cite{lingbot}, training on 2,500 Clean demonstrations (50 per task) combined with 25,000 Random demonstrations (500 per task) for up to 50k steps. We train on 64 GPUs and evaluate at multiple frame rates to study the effect of frame rate on training convergence and inference acceleration. For ablation studies, we train on 16 GPUs using only the 2,500 Clean demonstrations at 25\,fps for 20k steps for efficient comparison.

\subsection{Main Results}

\subsubsection{Comparison with State-of-the-Art}

Table~\ref{tab:benchmark} compares \method with state-of-the-art methods on the RoboTwin benchmark. \method achieves the highest success rate on both Clean (94.1\%) and Random (93.5\%) settings, outperforming all compared methods including both VLA and WAM approaches. 

\subsubsection{Training Convergence Analysis}

Figure~\ref{fig:teaser} presents a detailed comparison of training convergence between \method and LingBot-VA across two frame rates and two difficulty levels  (full numerical results in Appendix~\ref{app:convergence}).

At 12\,fps, \method converges roughly 2$\times$ faster than LingBot-VA, reaching 90.0\% at 10k steps versus the baseline's $\sim$20k steps, and achieves higher final accuracy (94.1/93.5\% vs.\ 92.8/91.8\% on Clean/Random). The advantage becomes dramatically more pronounced at 50\,fps: at 5k steps, \method already achieves 70.2/61.6\% versus LingBot-VA's 45.5/31.9\%, and maintains the lead through convergence. At 50k steps, \method reaches 91.8/90.5\% versus 88.6/85.2\%. The large gap at high frame rates confirms that multi-chunk prediction is particularly effective when adjacent frames are nearly identical, forcing the model to learn meaningful dynamics rather than relying on appearance shortcuts.

\paragraph{Why Does MCP Work Better at High Frame Rates?}              The dramatic improvement at 50\,fps can be understood through supervision signal density. At high frame rates, adjacent chunks are nearly identical, making next-chunk denoising trivially solvable via appearance copying. Multi-chunk prediction breaks this shortcut: chunks 2 or 3 steps ahead exhibit substantial visual differences that can only be predicted by understanding the underlying physical dynamics, forcing the model to develop temporally-aware representations.

\subsubsection{PhyWorld Benchmark}

\begin{figure}[t]
    \centering
    \includegraphics[width=1\linewidth]{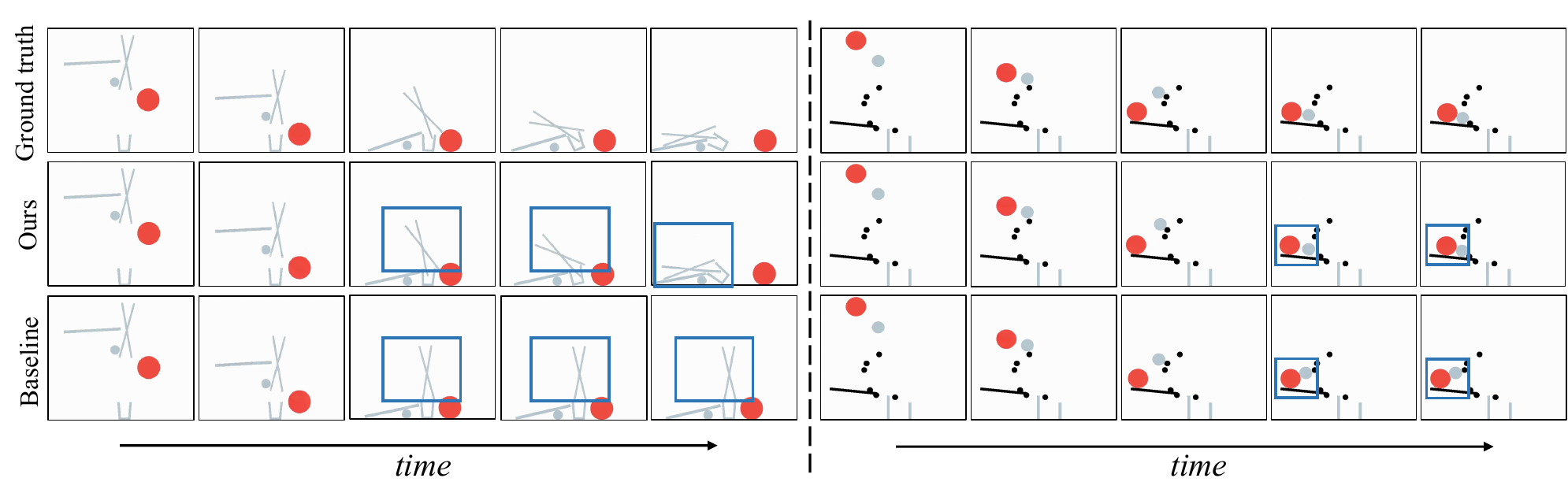}
    \caption{\textbf{Qualitative comparison on PhyWorld.} We show 5 frames (start, 3 intermediate, end) from ground truth (top), \method (middle), and Baseline (bottom). Blue boxes highlight regions where the baseline deviates from the ground-truth physical trajectory, while \method generates more physically consistent dynamics.}
    \label{fig:phyworld}
\end{figure}

To evaluate whether MCP improves temporal and physical dynamics modeling beyond robot manipulation, we additionally evaluate \method on PhyWorld~\cite{phyworld} in a video-only setting, removing the action stream from both \method and the baseline. We report Frechet Video Distance (FVD~\cite{fvd}, lower is better) and Abnormal Ratio (percentage of generated videos violating physical laws, lower is better).

\begin{table}[t]
  \caption{Evaluation on PhyWorld benchmark for combinatorial generalization. We report FVD and Abnormal Ratio under out-of-template (OOT) and in-template (IT) settings.}
  \label{tab:phyworld}
  \centering
  \small
  \begin{tabular}{@{}lcccc@{}}
    \toprule
    & \multicolumn{2}{c}{FVD ($\downarrow$)} & \multicolumn{2}{c}{Abnormal Ratio ($\downarrow$)} \\
    \cmidrule(lr){2-3} \cmidrule(lr){4-5}
    Method & OOT & IT & OOT & IT \\
    \midrule
    LingBot-VA & 5.3 & 3.5 & 12\% & 3\% \\
    \textbf{\method} & \textbf{4.7} & \textbf{3.2} & \textbf{8\%} & \textbf{2\%} \\
    \bottomrule
  \end{tabular}
\end{table}

As shown in Table~\ref{tab:phyworld} and Figure~\ref{fig:phyworld}, \method improves both FVD and Abnormal Ratio over LingBot-VA, with larger gains on the out-of-template setting, suggesting that multi-chunk prediction encourages generalizable physical dynamics rather than template-specific memorization.

\subsubsection{Generality on Video Pretraining}
\label{sec:general_video}

Beyond World Action Models, we further evaluate \method on action-free general video pretraining. We conduct pretraining experiments on an in-house general video dataset comprising approximately 3.5M video clips of 5--10 seconds, predominantly featuring human activities. We remove the action stream, train on 32 GPUs, and evaluate pure video generation quality using FVD~\cite{fvd} (lower is better). We construct two held-out test sets of 1,024 samples each: Test Set~1 contains human activity videos, while Test Set~2 focuses on camera-driven scene dynamics. 

\begin{figure}[t]
    \centering
    \includegraphics[width=0.75\linewidth]{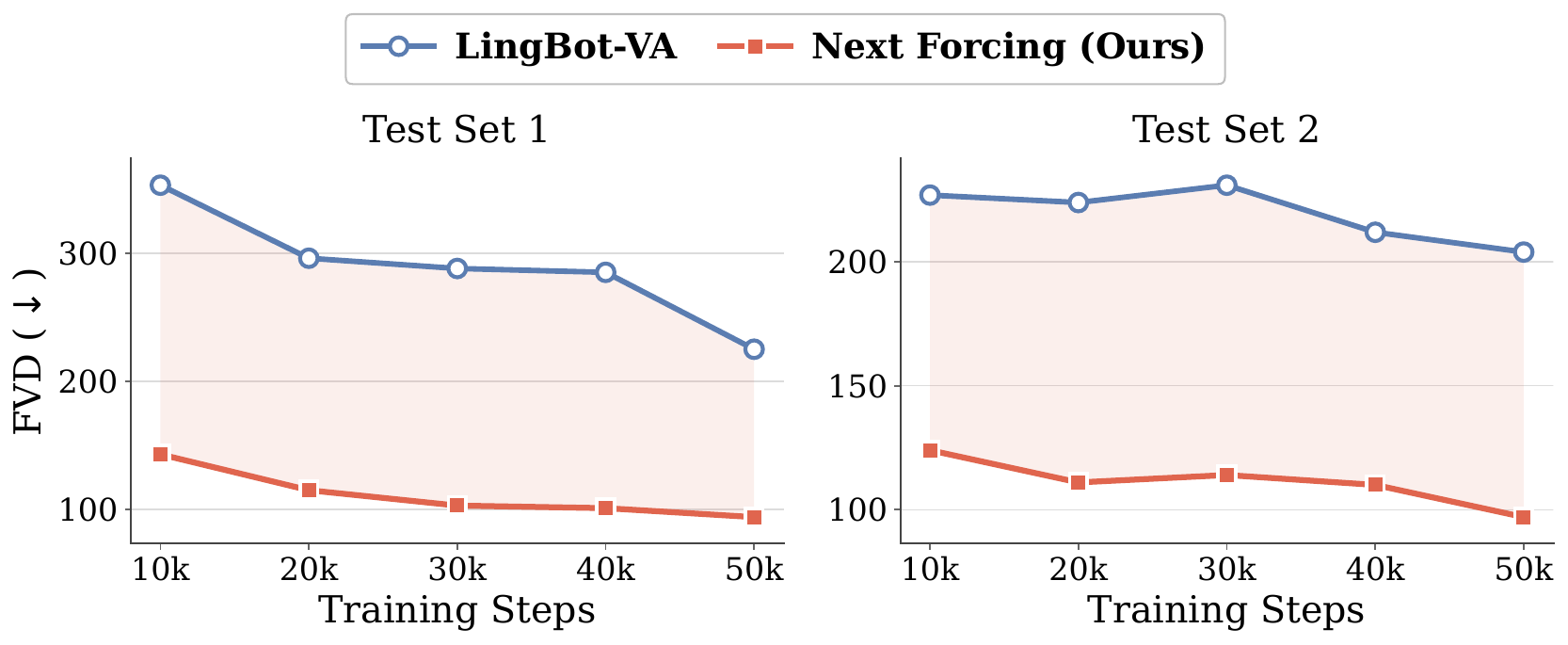}
    \caption{\textbf{FVD ($\downarrow$) on general video pretraining across training steps.} Test Set~1 contains human activity videos, while Test Set~2 focuses on camera-driven scene dynamics. \method consistently achieves substantially lower FVD than LingBot-VA on both test sets throughout training.}
    \label{fig:fvd_curves}
\end{figure}

As shown in Figure~\ref{fig:fvd_curves}, \method consistently achieves substantially lower FVD than LingBot-VA throughout training on both test sets. At 50k steps, \method reduces FVD by 58\% on Test Set~1 (94 \vs 225) and 52\% on Test Set~2 (97 \vs 204). Notably, \method at only 10k steps already surpasses LingBot-VA at 50k steps on both sets, demonstrating significantly faster convergence in video quality. These results confirm that multi-chunk prediction provides effective temporal supervision for general video generation, validating the generality of \method beyond the robot manipulation domain.

\subsection{Ablation Studies}

We conduct systematic ablation studies on both the baseline and MCP module design choices. All ablations are trained and evaluated on the RoboTwin Clean subset for 20k steps (Table~\ref{tab:ablation}).

\paragraph{Baseline Ablation.}

We first ablate key design choices of the baseline LingBot-VA to understand their impact. The default baseline uses $s_\text{main} = 5$ with noisy history augmentation, achieving 75.6\%. Removing noisy history augmentation causes a significant drop to 69.8\%, as the model can shortcut the denoising task by directly copying from the clean context rather than learning meaningful dynamics. The timestep shift has a substantial effect: $s_\text{main} = 1$ yields 65.3\%, $s_\text{main} = 10$ achieves the best at 78.4\%, while $s_\text{main} = 20/25$ gives 77.6/77.2\%, showing diminishing returns. Adding MCP improves performance from 75.6\% to 85.8\% (+10.2\%).

\paragraph{MCP Module Ablation.}

From the default MCP configuration (Table~\ref{tab:ablation}), we ablate each design choice. Removing multi-layer fusion drops to 83.6\%, confirming that fusing intermediate features enables deeper gradient propagation into the main model. Setting $s_\text{mcp} = 5$ (same as main) reduces to 83.2\%, as the higher shift forces the MCP modules to rely more on the main model's representations and strengthens their coupling. Without weight initialization yields 83.8\%. For transformer blocks, 1 block achieves 86.5\% and 5 blocks gives 85.0\%, suggesting that lighter MCP modules result in tighter coupling and more effective supervision. Despite the slightly higher rate with 1 block, we default to 3 blocks as it produces fewer visual artifacts in the MCP-generated chunks, which is important for the parallel chunk generation mode at inference.

\subsection{Inference Acceleration}




As described in Section~\ref{sec:inference}, the MCP modules can be retained to generate two video chunks per autoregressive step, reducing the number of sequential video denoising steps. Table~\ref{tab:inference} shows a frame-rate-dependent throughput--accuracy trade-off. At 12 and 25\,fps, MCP-accelerated inference reduces task success rate, with the largest drop being 2.9 percentage points on the 12\,fps Random setting. At 50\,fps, it matches or slightly exceeds standard inference in this evaluation. The accelerated mode is therefore more suitable at high frame rates, while standard inference can be used when strict accuracy parity is required.

\begin{table}[t]
  \caption{Inference acceleration. Standard inference performs separate video denoising for each chunk. MCP-accelerated inference predicts the next video chunk in parallel via the MCP module.}
  \label{tab:inference}
  \centering
  \small
  \begin{tabular}{@{}lcccccc@{}}
    \toprule
    & \multicolumn{2}{c}{12\,fps} & \multicolumn{2}{c}{25\,fps} & \multicolumn{2}{c}{50\,fps} \\
    \cmidrule(lr){2-3} \cmidrule(lr){4-5} \cmidrule(lr){6-7}
    Inference Mode & Clean & Random & Clean & Random & Clean & Random \\
    \midrule
    Standard & 94.1 & 93.5 & 92.6 & 91.4 & 91.8 & 90.5 \\
    MCP-accelerated (2$\times$) & 93.5 & 90.6 & 91.0 & 89.8 & 92.2 & 91.3 \\
    \bottomrule
  \end{tabular}
\end{table}

\section{Conclusion}
\label{conclusion}

We presented \textbf{\method}, a multi-chunk prediction framework for World Action Models that addresses myopic supervision in autoregressive video prediction. By improving future visual representations, \method also benefits action decoding in joint video-action architectures. By training lightweight MCP modules to predict multiple future chunks alongside the main model, \method provides dense temporal supervision that forces the model to learn long-range dynamics rather than relying on appearance shortcuts. \method establishes new state-of-the-art results on RoboTwin (94.1/93.5\% on Clean/Random), achieves 2.3$\times$ faster training convergence at 50\,fps, enables inference acceleration, and demonstrates over 50\% FVD reduction on general video pretraining, validating its generality beyond robot-specific data and applicability to action-free video generation. The main limitation is that the MCP modules introduce extra training cost. We hope \method motivates further investigation of training objectives, beyond context construction and noise scheduling, as a key axis for improving autoregressive video generation.

\begin{ack}
This research is supported by the National Natural Science Foundation of China (623B2036, U25B2045, 62472184), the Fundamental Research Funds for the Central Universities, and the Innovation Project of Optics Valley Laboratory (Grant No.\ OVL2025YZ005).
\end{ack}

\clearpage
{
\bibliographystyle{plain}
\bibliography{reference}
}








\appendix
\newpage
\appendix

\section*{Supplementary Material}

This appendix provides additional details that complement the main paper. Appendix~\ref{app:attn_mask} describes the attention mask design shared by the main model and MCP modules. Appendix~\ref{app:convergence} presents the full numerical results for the training convergence experiments. Appendix~\ref{app:timestep_shift} details the timestep shift formulation used for noise scheduling.

\section{Attention Mask Details}
\label{app:attn_mask}

\begin{figure}[h]
    \centering
    \includegraphics[width=1\linewidth]{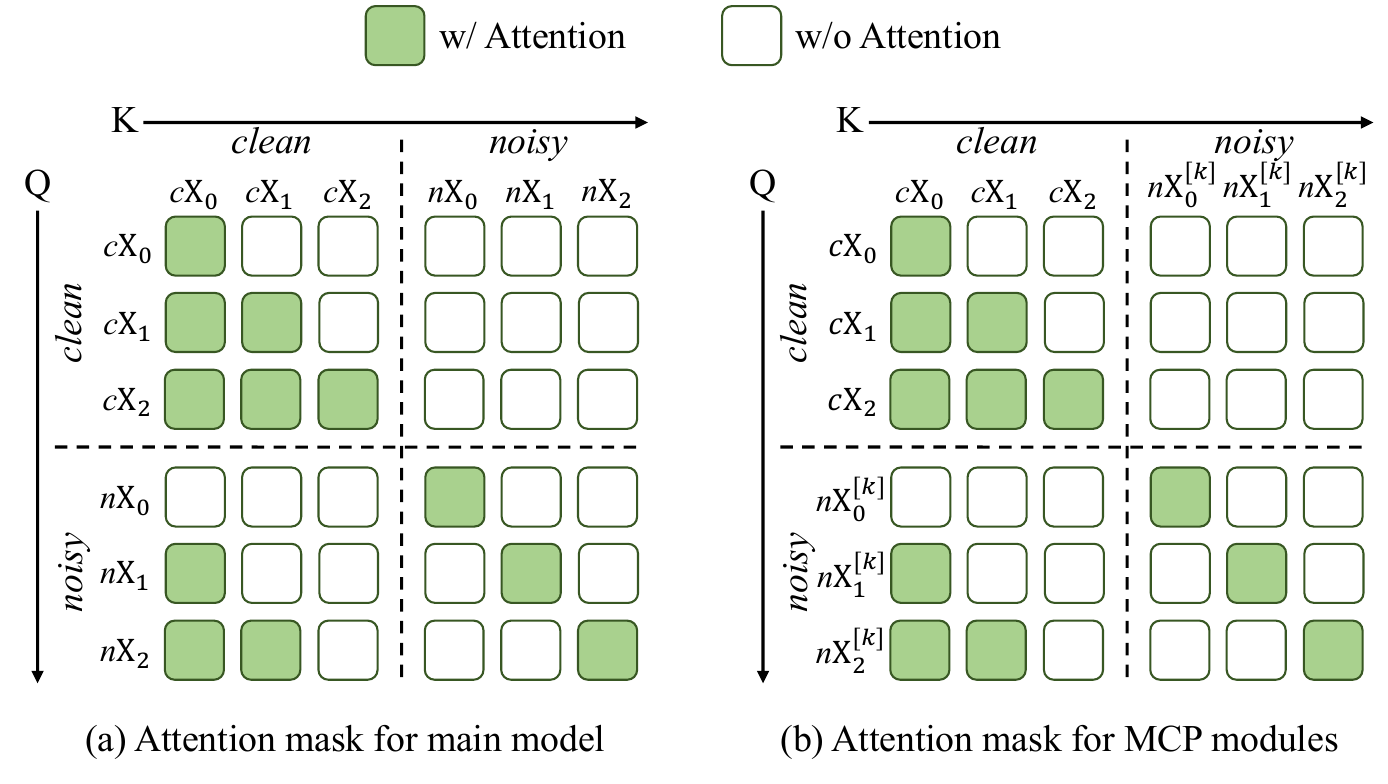}
    \caption{\textbf{Attention mask for main model and MCP modules.} Only video tokens are shown for clarity (action tokens omitted). Under teacher forcing, the sequence consists of noisy tokens (current chunk being denoised) and clean tokens (ground-truth context). Noisy tokens attend to all causally preceding clean tokens and to noisy tokens within the same chunk; clean tokens follow a standard causal pattern; clean tokens cannot attend to noisy tokens. The MCP modules share the same attention mask structure as the main model, requiring only a single mask construction per training step.}
    \label{fig:attn_mask}
\end{figure}

Figure~\ref{fig:attn_mask} illustrates the attention mask shared by both the main model and MCP modules. For clarity, only video tokens are shown (action tokens omitted). The sequence consists of two groups: noisy tokens (the current chunk being denoised) and clean tokens (ground-truth context from previous chunks). The attention rules under teacher forcing are:

\begin{itemize}
    \item \textbf{Noisy $\rightarrow$ Clean:} Each noisy token attends to all causally preceding clean tokens, excluding the clean token at the same chunk index. This prevents information leakage from the current chunk's ground truth.
    \item \textbf{Noisy $\rightarrow$ Noisy:} Noisy tokens only attend to other noisy tokens within the same chunk (self-attention within the chunk being denoised).
    \item \textbf{Clean $\rightarrow$ Clean:} Clean tokens follow a standard causal pattern, attending to all clean tokens at the same or earlier chunk indices.
    \item \textbf{Clean $\rightarrow$ Noisy:} Not permitted. Clean context tokens cannot attend to noisy tokens.
\end{itemize}

A key design choice of \method is that the MCP modules reuse the same attention mask as the main model. Since both operate on sequences with the same structure (noisy target tokens + clean context tokens), the mask can be constructed once per training step and shared across the main model and all MCP depths, reducing training overhead.

\section{Detailed Training Convergence Results}
\label{app:convergence}

Table~\ref{tab:convergence_detail} provides the full training convergence data at 12\,fps and 50\,fps, complementing the convergence curves in Figure~\ref{fig:teaser}. Success rates are reported every 5k training steps from 5k to 50k.

\begin{table}[h]
  \caption{Task success rate (\%) on RoboTwin across training steps at 12\,fps and 50\,fps. \method consistently outperforms LingBot-VA. At 12\,fps, \method leads by $\sim$10 points at 5k steps and maintains a 1--2 point advantage at convergence. At 50\,fps, the gap is substantially larger: \method leads by 24.7/29.7 points (Clean/Random) at 5k steps, and the advantage persists through convergence.}
  \label{tab:convergence_detail}
  \centering
  \small
  \begin{tabular}{@{}cclcccccccccc@{}}
    \toprule
    FPS & Setting & Method & 5k & 10k & 15k & 20k & 25k & 30k & 35k & 40k & 45k & 50k \\
    \midrule
    \multirow{4}{*}{12}
    & \multirow{2}{*}{Clean}
    & LingBot-VA & 74.0 & 85.2 & 87.8 & 90.8 & 92.3 & 91.3 & 92.8 & 92.9 & 93.1 & 92.8 \\
    & & \method & \textbf{84.9} & \textbf{90.0} & \textbf{91.5} & \textbf{92.3} & \textbf{93.3} & \textbf{94.3} & \textbf{93.1} & \textbf{93.4} & \textbf{94.2} & \textbf{94.1} \\
    \cmidrule(l){2-13}
    & \multirow{2}{*}{Random}
    & LingBot-VA & 73.5 & 82.2 & 85.0 & 88.3 & 88.9 & 89.4 & 89.3 & 91.2 & 91.4 & 91.8 \\
    & & \method & \textbf{80.6} & \textbf{85.4} & \textbf{85.8} & \textbf{90.5} & \textbf{89.8} & \textbf{91.5} & \textbf{89.6} & \textbf{91.5} & \textbf{91.6} & \textbf{93.5} \\
    \midrule
    \multirow{4}{*}{50}
    & \multirow{2}{*}{Clean}
    & LingBot-VA & 45.5 & 64.8 & 69.6 & 78.5 & 79.0 & 81.2 & 82.4 & 83.8 & 87.4 & 88.6 \\
    & & \method & \textbf{70.2} & \textbf{80.5} & \textbf{85.2} & \textbf{87.4} & \textbf{87.6} & \textbf{90.0} & \textbf{90.9} & \textbf{91.5} & \textbf{91.7} & \textbf{91.8} \\
    \cmidrule(l){2-13}
    & \multirow{2}{*}{Random}
    & LingBot-VA & 31.9 & 54.7 & 59.8 & 69.4 & 70.7 & 75.6 & 79.2 & 80.4 & 84.5 & 85.2 \\
    & & \method & \textbf{61.6} & \textbf{77.6} & \textbf{80.2} & \textbf{85.0} & \textbf{85.4} & \textbf{86.8} & \textbf{89.9} & \textbf{88.4} & \textbf{89.6} & \textbf{90.5} \\
    \bottomrule
  \end{tabular}
\end{table}

Several observations emerge from the detailed results. First, \method shows the largest absolute improvement in the early training stages (5k--10k steps), indicating that multi-chunk prediction provides a stronger learning signal from the start by preventing appearance shortcuts. Second, the improvement from MCP is notably frame-rate dependent. At 12\,fps, adjacent chunks differ substantially in visual content, so the appearance shortcut is less severe and the baseline can still learn meaningful dynamics, resulting in a moderate improvement from MCP. At 50\,fps, however, adjacent chunks are nearly identical, making it easy for the model to shortcut by simply copying the previous chunk. In this regime, the baseline struggles to learn beyond local copying, while MCP forces the model to capture long-range dynamics, leading to dramatically larger gains.

\section{Timestep Shift Formulation}
\label{app:timestep_shift}

Here we provide the detailed formulation of the timestep shift mechanism used in both the main model and MCP modules.

\paragraph{Shifted Timestep Schedule.}
We first construct a schedule of $T = 1000$ uniformly spaced base values $\{\sigma_i\}_{i=0}^{T-1}$ over $[\sigma_\text{min}, \sigma_\text{max}] = [0, 1]$. These are then transformed by the shift parameter $s$~\cite{sd3}:
\begin{equation}
    \tilde{\sigma}_i = \frac{s \cdot \sigma_i}{1 + (s - 1) \cdot \sigma_i}.
    \label{eq:shift_transform}
\end{equation}
This transformation is monotonic and maps $[0, 1] \to [0, 1]$, but redistributes the noise levels across the interval: a larger $s$ pushes $\tilde{\sigma}$ toward higher values, concentrating training on noisier regimes. For the main model we use $s_\text{main} = 5$; for the MCP modules we use $s_\text{mcp} = 10$.

\paragraph{Timestep Sampling.}
During training, a timestep index is uniformly sampled: $\text{id} \sim \text{Uniform}\{0, 1, \ldots, T-1\}$, and the corresponding shifted noise level $\tilde{\sigma}_\text{id}$ is looked up from the pre-computed schedule. The noisy training sample is then constructed as:
\begin{equation}
    \mathbf{x}_{\tilde{\sigma}} = (1 - \tilde{\sigma}_\text{id}) \, \mathbf{x}_0 + \tilde{\sigma}_\text{id} \, \boldsymbol{\epsilon}, \quad \boldsymbol{\epsilon} \sim \mathcal{N}(0, \mathbf{I}).
\end{equation}
Since the shift is already encoded in the schedule, uniform sampling over the shifted schedule effectively produces a non-uniform distribution over noise levels, biased toward higher noise for larger shift $s$.



\end{document}